\documentclass[letterpaper]{article} 
\usepackage{aaai2026}  
\usepackage{times}  
\usepackage{helvet}  
\usepackage{courier}  
\usepackage[hyphens]{url}  
\usepackage{graphicx} 
\usepackage{natbib}  
\usepackage{caption} 
\usepackage{algorithm}
\usepackage{algorithmic}
\usepackage{pdfpages}

\usepackage{newfloat}
\usepackage{listings}
\DeclareCaptionStyle{ruled}{labelfont=normalfont,labelsep=colon,strut=off} 
\floatstyle{ruled}
\newfloat{listing}{tb}{lst}{}
\floatname{listing}{Listing}

\usepackage{amsmath} 
\usepackage{booktabs}
\usepackage{amssymb}

\newcommand\mc{\mathcal}
\nocopyright 

\title{Aligning Medical Conversational AI through Online Reinforcement Learning with Information-Theoretic Rewards}
\author{
    Tanvi Verma,
    Yang Zhou,
    Rick Siow Mong Goh,
    Yong Liu
}
\affiliations{

    Institute of High Performance Computing, Agency for Science, Technology and Research (A*STAR), Singapore \\ tanviverma.2015@phdcs.smu.edu.sg

}

\begin{document}

\maketitle

\begin{abstract}
We present Information Gain Fine-Tuning (IGFT), a novel approach for training medical conversational AI to conduct effective patient interviews and generate comprehensive History of Present Illness (HPI) without requiring pre-collected human conversations. IGFT combines online Group Relative Policy Optimization (GRPO) with information-theoretic rewards, enabling models to learn from self-generated conversations with simulated patients. Unlike existing approaches that rely on expensive expert-annotated conversations or static datasets, our online RL framework allows models to discover effective questioning strategies through exploration.
Our key innovation is an information gain reward function that tracks which clinical entities such as symptoms, temporal patterns, and medical history, are revealed during conversation. Each question's reward is computed based on its expected information gain combined with GPT-4o-mini quality assessments across dimensions including clinical relevance, patient engagement, and specificity. This hybrid approach ensures models learn to ask targeted, clinically appropriate questions that efficiently gather diagnostic information.
We fine-tune two models using LoRA: Llama-3.1-8B-Instruct and DeepSeek-R1-Distill-Qwen-7B (a reasoning-optimized model). Training exclusively on Avey data containing concise HPIs, we evaluate generalization to MIMIC data with longer, more elaborate HPIs. DeepSeek-R1-Distill-Qwen-7B (IGFT) achieves F1 scores of 0.408 on Avey (10.9\% improvement over base) and 0.289 on MIMIC (12.9\% improvement), while Llama-3.1-8B-Instruct (IGFT) reaches 0.384 and 0.336 respectively. Both models outperform OpenAI's model on MIMIC and surpass medical domain-specific baselines like HuatuoGPT and UltraMedical, which were optimized for single-turn medical QA rather than multi-turn conversations.
These results demonstrate that IGFT enables models to learn effective questioning strategies without human supervision, offering a scalable approach for training AI assistants to support healthcare providers in clinical history-taking.
\end{abstract}

\section{Introduction}

The integration of artificial intelligence into healthcare workflows remains an active area of research with both promising applications and significant challenges. While there is growing interest in AI-assisted healthcare, public and professional trust in AI for medical tasks is still developing, particularly for critical decisions like diagnosis. However, there may be opportunities for AI to assist with specific, well-defined tasks that could support healthcare providers. One such task is the collection of patient histories, a fundamental yet time-consuming component of medical consultations that often accounts for approximately 20--30\% of the clinical encounter time \cite{wei2022physical, desai2021automated}. An AI assistant capable of conducting preliminary patient interviews to generate History of Present Illness (HPI) documentation could potentially support clinical workflows, though such systems would require careful validation and human oversight.

Training AI systems for medical history-taking presents unique alignment challenges. The AI must learn to ask clinically relevant questions that systematically uncover patient information while maintaining appropriate boundaries and avoiding harmful or irrelevant queries. Current approaches to developing such systems face several limitations:

\noindent \textbf{Data Availability Challenges:} Supervised learning approaches require extensive datasets of expert conversations. These are resource-intensive to collect, may not represent the full diversity of patient presentations, and often reflect individual practitioner styles rather than standardized best practices. The scarcity of high-quality medical conversation data remains a significant bottleneck.

\noindent \textbf{Limited Exploration:} Existing methods, whether supervised or based on offline reinforcement learning, constrain models to patterns present in their training data. This prevents the discovery of potentially more efficient questioning strategies and limits adaptability to novel situations.

\noindent \textbf{Objective Definition:} More fundamentally, learning from human demonstrations assumes we know the optimal way to conduct medical interviews. However, even experienced clinicians may have different approaches, and what constitutes ``optimal'' information gathering remains an open question.

We propose an alternative approach: rather than training AI to imitate human doctors, we train it to optimize for information gain through online reinforcement learning (RL). In the RL context, ``online'' means the training data is generated from the same policy we are optimizing---the model learns from conversations created by its current policy, receives rewards, and updates itself iteratively. This contrasts with offline RL, where the model learns from conversations generated by some other policy (e.g., human doctors or a previous model version) without the ability to explore new strategies. Our online approach enables the model to discover effective questioning strategies through exploration: it generates questions, observes their information gain through patient responses, and adjusts its policy to ask better questions in future iterations.

Our key insight is that medical history-taking can be formalized as an information acquisition problem. Each question should reduce uncertainty about the patient's condition by revealing previously unknown symptoms, temporal patterns, or relevant medical history. By tracking which clinical entities have been uncovered during a conversation and computing the expected information gain of potential questions, we can provide a principled reward signal for learning.

To promote clinical appropriateness, we augment our information-theoretic rewards with quality assessments from GPT-4o-mini, evaluating factors such as clinical relevance, patient engagement potential, and question specificity. This hybrid approach aims to balance information gathering efficiency with adherence to medical communication norms.

We implement this approach using Group Relative Policy Optimization (GRPO), a variant of policy gradient methods designed for language model fine-tuning. By comparing multiple generated questions and optimizing toward those with higher information gain, GRPO enables stable learning from our reward signal. We evaluate our method, which we call Information Gain Fine-Tuning (IGFT), on two models: Llama-3.1-8B-Instruct, a general-purpose instruction-following model, and DeepSeek-R1-Distill-Qwen-7B, a model optimized for reasoning tasks. This comparison allows us to investigate whether reasoning capabilities influence strategic questioning abilities.

Our contributions are:
\begin{itemize}
\item \textbf{An online learning framework} for medical dialogue systems that learns from self-play with simulated patients, reducing dependency on scarce expert-annotated data.
\item \textbf{An information-theoretic reward function} that formalizes medical history-taking as entropy minimization over clinical entities, providing a principled training objective.
\item \textbf{Safety considerations} through LLM-based quality assessment, aiming to ensure that information maximization does not lead to inappropriate questions.
\item \textbf{Empirical evaluation} showing that online GRPO with information gain rewards can train models that achieve improved precision, recall, and F1 scores in HPI entity extraction compared to base models.
\item \textbf{A methodology} for exploring AI alignment in specialized domains through well-defined objectives rather than extensive human supervision.
\end{itemize}

This work investigates whether information-theoretic principles can guide the development of AI systems for structured tasks like medical history-taking. While such systems would require extensive validation before clinical deployment, our results suggest that principled reward design may offer a path toward AI tools that could eventually support healthcare providers in specific, well-defined tasks under appropriate human oversight.

\section{Related Work}

\subsection{Medical Dialogue Systems}

Medical dialogue systems have been comprehensively surveyed in \cite{shi2024medical}, which categorizes work across rule-based systems, neural architectures, evaluation practices, and grand challenges specific to medical LLMs. The survey reveals a critical limitation: many existing medical LLMs operate in a single-turn, passive QA paradigm, assuming patients provide comprehensive symptom descriptions up front \cite{tu2024towards}. Models such as MedAlpaca \cite{han2023medalpaca} and BioMistral \cite{labrak2024biomistral} focus on direct answer generation from static prompts, lacking mechanisms for strategic, multi-turn symptom elicitation. While resources like MedDialog \cite{zeng-etal-2020-meddialog} provide multi-turn patient–doctor conversations, models trained on them often default to generic responses without adaptive reasoning. 

Recent systems have attempted to address some of these limitations through different approaches. ChatDoctor \cite{li2023chatdoctor} fine-tunes LLaMA-7B and incorporates retrieval-augmented generation to improve response accuracy. ClinicalGPT \cite{wang2023clinicalgpt} specifically targets multi-round clinical tasks, training on medical records and patient dialogues to enhance diagnostic reasoning capabilities. However, both systems still fundamentally rely on learning from existing conversation patterns in their training data.

Reinforcement learning approaches have shown promise for medical dialogue systems, including early MDP-based methods \cite{wei2018task, xu2019end}, hierarchical RL \cite{zhong2022hierarchical}, and the recent DoctorAgent-RL \cite{feng2025doctoragent} which uses multi-agent collaboration for clinical dialogue. However, these RL approaches operate in offline settings, learning from fixed datasets rather than through online exploration.

Similarly, knowledge-enhanced models such as HuaTuoGPT \cite{wang2304huatuo} and UltraMedical \cite{zhang2024ultramedical} have limited evidence of engaging in targeted, information-seeking dialogue. These approaches generally treat medical conversations as sequence generation tasks, without modeling the strategic decision-making needed in ambiguous or multi-symptom cases.

In contrast, our method frames medical interviewing as an online, reinforcement-learning process driven by information gain, enabling adaptive, question-driven interactions without requiring human-annotated conversations. This directly addresses what \cite{shi2024medical} identify as key limitations: the reliance on static datasets and the inability to explore novel questioning strategies beyond those present in training data.

\subsection{Information-Theoretic Approaches in AI}

Information theory has supplied principled foundations for AI tasks requiring strategic information gathering. In active learning, information gain (e.g., entropy reduction) is a widely used criterion for selecting data samples to minimize annotation costs while maximizing learning efficiency \cite{settles2009active}.

In question generation, mutual information has been employed to steer models toward generating highly informative, goal-oriented questions. Visual question generation systems like \cite{krishna2019information} explicitly model mutual information between the question, underlying answer categories, and image content to avoid generic questions .

In RL, curiosity-driven approaches such as VIME (Variational Information Maximizing Exploration) explicitly reward information gain, measuring how actions update the agent's belief over environment dynamics, to encourage exploration in continuous state spaces \cite{houthooft2016vime}.

Information-theoretic methods have also been applied in dialogue and negotiation AI, where strategic questioning enables agents to uncover hidden preferences. In negotiation research, agents learn to probe opponents via sequential interactions guided by simulated rollouts and expected information gains \cite{lewis2017deal}. In task-oriented dialogue, researchers have applied entropy minimization to reduce uncertainty over user goals and mutual information maximization to guide question selection \cite{lee2018answerer}.

However, applying these strategies to medical history-taking poses unique challenges: beyond maximizing information gain, systems must ensure clinical appropriateness, patient engagement, and diagnostic relevance. Our IGFT framework handles this by combining information-theoretic rewards with LLM-based question-quality assessment, ensuring questions are not only informative but also safe and clinically valid.

\subsection{Online Reinforcement Learning for Language Models}

While RL from human feedback \cite{christiano2017deep} has become the standard for aligning LLMs with static human preferences, recent work has explored online reinforcement learning to allow models to improve from self-generated interactions without human-labeled data.
Several methods have explored online reinforcement learning for language models using self-generated interaction loops rather than static human-labeled data. Notable examples include: Reinforced Self-Training (ReST), which iteratively generates sample responses and applies offline RL to refine the policy without human supervision \cite{gulcehre2023reinforced}; Self-Play Fine-Tuning (SPIN), where an LLM generates and compares multiple responses, learning through preference ranking self-play iterations \cite{chen2024self}; Reflect-RL, a two-player framework with a frozen reflection model that produces errors to train the policy via online RL \cite{zhou2024reflect}; and SCoRe, a multi-turn online RL approach enabling self-correcting behavior in language models without external annotation \cite{kumar2024training}. 

While these approaches demonstrate the effectiveness of online RL for improving LLM capabilities, they are not tailored for domain-specific applications such as medical dialogue. Our IGFT method uniquely combines information-theoretic rewards and automated LLM-based quality assessments to train models that conduct safe, targeted, and diagnostically relevant medical interviews from scratch.

\section{Information Gain Fine-Tuning (IGFT)}

\begin{figure*}[t] 
    \centering
    \includegraphics[width=\textwidth]{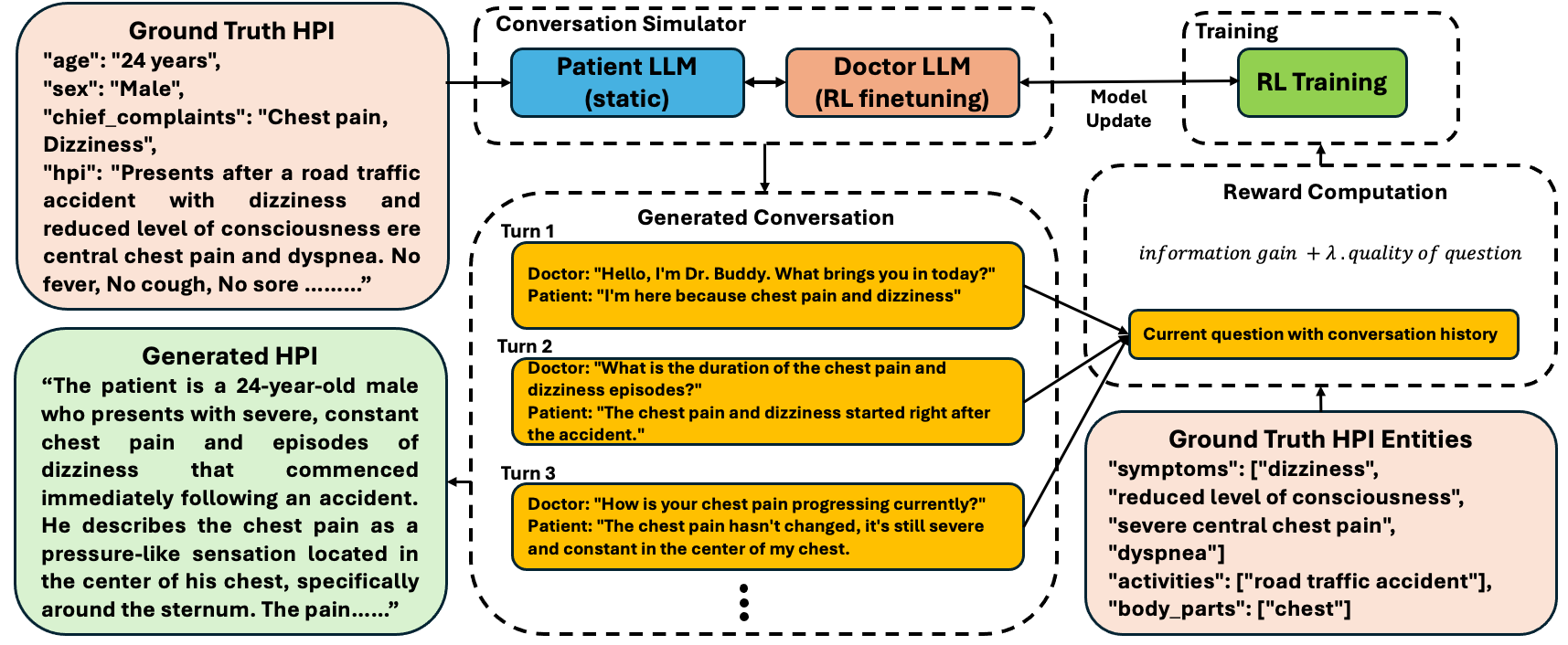}
    \caption{Online training framework for medical dialogue. The Doctor LLM generates questions answered by a Patient LLM simulating cases from ground truth HPIs. Information gain rewards are computed by comparing revealed entities against ground truth, combined with LLM-based quality assessment. The model updates through GRPO based on these rewards, creating an online learning loop that improves questioning strategies without requiring human conversation data. The Generated HPI (shown in green) is produced only during evaluation to measure performance, not during training.}
    \label{fig:fraemwork}
\end{figure*}

We present Information Gain Fine-Tuning (IGFT), a novel approach that combines online reinforcement learning with information-theoretic rewards to train medical conversational AI systems. Our method enables models to learn effective questioning strategies through self-play with simulated patients, without requiring human-annotated conversation data.

\subsection{Problem Formulation}

We model medical history-taking as a sequential decision-making problem where the goal is to maximize information gain about the patient's condition through strategic questioning. Formally, we define:

\begin{itemize}
    \item \textbf{State} $s_t$: The conversation history up to turn $t$, including all previous doctor questions and patient responses
    \item \textbf{Action} $a_t$: The doctor's question at turn $t$, generated by the policy $\pi_\theta$
    \item \textbf{Reward} $r_t$: The information gain achieved by asking question $a_t$ given the current state
\end{itemize}

During data collection, we generate complete N-turn doctor-patient conversations (episodes). From each episode, we extract multiple training samples, one for each turn, where each sample contains the conversation history up to that turn as the state. This provides diverse states with varying conversation lengths for training. The objective is to learn a policy $\pi_\theta$ that maximizes the expected cumulative information gain:

\begin{equation}
J(\theta) = \mathbb{E}_{\tau \sim \pi_\theta} \left[ \sum_{t=0}^{T} \gamma^t r_t \right]
\end{equation}

where $\tau$ represents a conversation trajectory and $\gamma$ is a discount factor (set to 1.0 in our experiments as conversations are relatively short).

\subsection{Information Gain Reward Function}

Our core innovation lies in formalizing the reward function based on information-theoretic principles. Medical history-taking can be viewed as a sequential information acquisition process, where each question aims to reduce uncertainty over the patient's condition. 

\subsubsection{Entity Coverage Tracking}
At each turn $t$, we maintain:
\begin{itemize}
    \item $\mathcal{E}$: The complete set of clinical entities from the patient's HPI
    \item $\mathcal{C}_t$: The set of entities covered (revealed) by turn $t$
    \item $\mathcal{U}_t = \mathcal{E} \setminus \mathcal{C}_t$: The set of uncovered entities remaining
\end{itemize}

This tracking mechanism allows us to compute information gain based on the actual progress of the conversation. As entities are revealed through patient responses, they move from $\mathcal{U}_t$ to $\mathcal{C}_t$, reducing the remaining uncertainty.

\subsubsection{Theoretical Foundation}
This uncertainty can be modeled using entropy, a foundational concept in information theory that quantifies the expected unpredictability of a random variable \cite{shannon1948mathematical}.

Given a discrete variable $X$ with distribution $p(x)$, the entropy is defined as:
\begin{equation}
H(X) = -\sum_{x \in \mathcal{X}} p(x) \log_2 p(x)
\end{equation}

In our setting, each unrevealed clinical entity $e \in \mc{U}_t$ (such as a symptom or historical fact) is treated as a binary random variable: it is either revealed during the conversation or not. Under maximum uncertainty (i.e., $p = 0.5$), the entropy per entity is maximal:
\begin{equation}
H_{\text{binary}}(0.5) = -0.5 \log_2 0.5 - 0.5 \log_2 0.5 = 1.0
\end{equation}

The entropy before asking a question $a_t$ is simply:
\begin{equation}
H(\mc{U}_t) = |\mc{U}_t| \cdot H_{\text{binary}}(0.5) = |\mc{U}_t| \cdot 1.0
\end{equation}

After asking $a_t$, we estimate for each $e \in \mc{U}_t$ the probability $p(e \mid a_t)$ that this question will elicit a response revealing $e$. The expected conditional entropy is:
\begin{equation}
H(\mc{U}_t \mid a_t) = \sum_{e \in \mc{U}_t} H_{\text{binary}}(p(e \mid a_t))
\end{equation}

The \textit{information gain} from asking question $a_t$ is then the reduction in entropy:
\begin{equation}
IG(a_t) = H(\mc{U}_t) - H(\mc{U}_t \mid a_t)
\end{equation}

This reward formulation encourages the model to ask questions that are most likely to uncover novel, relevant clinical information, driving it toward efficient and targeted information gathering.

\subsubsection{Entity Coverage Detection}

To determine which entities move from $\mathcal{U}_t$ to $\mathcal{C}_{t+1}$, we employ a multi-method approach:

\begin{enumerate}
    \item Exact phrase matching: Direct string matching with word boundaries
    \item Multi-word entity matching: All important words must be present
    \item Semantic similarity: Using sentence transformers with a similarity threshold $> 0.85$
\end{enumerate}
An entity is considered revealed if any of these methods indicates a match in the patient's response.

\subsubsection{Coverage Probability Estimation}
Since our online RL framework requires the model to select questions before observing patient responses, we must estimate the \textit{expected} information gain of each potential question. This requires predicting the probability $p(e|a_t)$ that asking question $a_t$ will reveal entity $e$.
We compute $p(e|a_t)$ using a weighted combination of complementary signals:

\begin{equation}
p(e|a_t) = \alpha \cdot \text{sem}(e, a_t) + \beta \cdot \text{llm}(e, a_t) + \gamma \cdot \text{key}(e, a_t)
\end{equation}

where:
\begin{itemize}
    \item $\text{sem}(e, a_t)$: Semantic similarity between entity and question using sentence embeddings
    \item $\text{llm}(e, a_t)$: LLM-based relevance score from GPT-4o-mini assessment
    \item $\text{key}(e, a_t)$: Keyword matching score
    \item $\alpha, \beta, \gamma$: Weighting parameters adding to 1.0
\end{itemize}
This multi-signal approach is crucial because different question types reveal information differently. Direct questions (``Do you have fever?'') benefit from keyword matching, while indirect questions (``How has your condition changed since yesterday?'') rely more on semantic understanding to predict they might reveal temporal patterns.
We clip probabilities to $[0.05, 0.95]$ to avoid degenerate cases where entropy becomes zero.

\subsubsection{Clinical Importance Weighting}

Different entity categories have varying clinical importance. We apply category-specific weights:

\begin{equation}
IG_{weighted}(a_t) = \sum_{c \in \text{categories}} w_c \cdot IG_c(a_t) \label{eq:gain}
\end{equation}

where $w_c$ represents the clinical importance weight for category $c$.

\subsection{LLM-Based Quality Assessment}

Pure information gain could lead to inappropriate or clinically irrelevant questions. We augment our reward with quality assessments from GPT-4o-mini across five dimensions:

\begin{enumerate}
    \item \textbf{Information gathering} (0-1): Likelihood of eliciting new, relevant information
    \item \textbf{Specificity} (0-1): How well the question targets uncovered elements
    \item \textbf{Patient engagement} (0-1): Expected quality of patient response
    \item \textbf{Clinical relevance} (0-1): Relevance to presenting concerns
    \item \textbf{Comprehensiveness} (0-1): Ability to uncover multiple aspects
\end{enumerate}

The final reward combines information gain with quality bonuses:

\begin{equation}
r_t = IG_{weighted}(a_t) + \lambda \cdot \text{quality}(a_t) \label{eq:reward}
\end{equation}

where $\lambda$ balances information gain with conversation quality.

\subsection{Online Training with GRPO}

We employ Group Relative Policy Optimization (GRPO) \cite{shao2024deepseekmath} for stable policy learning. GRPO is a ranking-based reinforcement learning method tailored for fine-tuning language models. Unlike traditional policy gradient methods that rely on absolute reward values, GRPO compares a group of candidate actions (e.g., model-generated questions) and optimizes the policy toward those with relatively higher rewards. This ranking-based formulation improves training stability and sample efficiency, particularly in language generation tasks where reward signals may be sparse or noisy. GRPO generates multiple responses per prompt and updates the model using a softmax-weighted ranking objective that favors higher-reward outputs.

\begin{algorithm}[h]
\caption{GRPO with Information Gain Rewards}
\label{alg:grpo}
\begin{algorithmic}[1]
\REQUIRE Dataset $\mathcal{D}$, base model $\pi_{\theta_0}$, reward function $R$
\ENSURE Optimized policy $\pi_{\theta^*}$
\FOR{each epoch}
    \FOR{each batch of conversations}
        \STATE Extract uncovered entities $U_t$ for each conversation
        \STATE Generate $K$ questions $\{a_i\}_{i=1}^K \sim \pi_\theta(a|s_t)$
        \FOR{each question $a_i$}
            \STATE Compute $IG(a_i)$ using Eq. \ref{eq:gain}
            \STATE Assess quality via LLM
            \STATE Calculate reward $r_i$ using Eq. \ref{eq:reward}
        \ENDFOR
        \STATE Update $\theta$ using GRPO loss (Eq. \ref{eq:loss})
    \ENDFOR
\ENDFOR
\RETURN $\pi_{\theta^*}$
\end{algorithmic}
\end{algorithm}

The GRPO loss function is:

\begin{equation}
\mathcal{L}(\theta) = -\mathbb{E}_{a \sim \pi_\theta} \left[ \sum_{i=1}^{K} \frac{\exp(r_i/\tau)}{\sum_j \exp(r_j/\tau)} \log \pi_\theta(a_i|s) \right] \label{eq:loss}
\end{equation}

where $\tau$ is a temperature parameter controlling the sharpness of the ranking. We apply LoRA (Low-Rank Adaptation) \cite{hu2022lora} to efficiently fine-tune large language models. 

\subsection{Training Procedure}

Our training procedure, illustrated in Figure \ref{fig:fraemwork}, follows an online reinforcement learning loop where the model learns from self-generated conversations:
\begin{enumerate}
    \item \textbf{Patient Simulation}: For each patient in the training set, we initialize a simulated patient using GPT-4o-mini with the ground truth HPI and instructions to reveal information gradually
    \item \textbf{Conversation Generation}: The doctor model generates questions, the patient responds, creating conversation trajectories
    \item \textbf{Reward Computation}: For each doctor question, we compute information gain based on which entities it helps uncover
    \item \textbf{Policy Update}: Update the doctor model using GRPO to favor high-information-gain questions
    \item \textbf{Iteration}: Repeat for multiple epochs, cycling through different patients
\end{enumerate}
Figure \ref{fig:fraemwork} visualizes this online learning framework, showing how the Doctor LLM and Patient LLM interact to generate training data, how rewards are computed based on revealed entities, and how the model updates through GRPO. The key insight is that unlike traditional supervised approaches that require pre-collected expert conversations, our method generates its own training data through this self-play mechanism.
\section{Experimental Setup}

\subsection{Dataset}
\subsubsection{Training:} We finetune on 350 vignettes from the Avey AI Benchmark vignette suite\cite{avey2024benchmark, hammoud2022avey}, which contains patient vignettes created by a team of doctors using medical websites and materials. The vignettes include patient demographics (age, sex), chief complaints, history of present illness, absent findings, physical examination notes, past medical/surgical/family history, and differential diagnoses. As noted in our methodology, we use pre-extracted clinical entities across 10 categories, with each case containing an average of 10-15 entities.

\subsubsection{Testing:} We test on 48 held-out HPIs from Avey and 50 HPIs from MIMIC-IV dataset \cite{johnson2020mimic}. Cases from MIMIC-IV contain longer, more elaborate HPIs, providing a challenging generalization test since models were trained only on Avey's concise vignettes. 

This setup tests whether IGFT learns questioning strategies that generalize beyond the specific format and style of the Avey training data.

\subsection{Models and Baselines}
We evaluate our approach on two base models: Llama-3.1-8B-Instruct \cite{llama2024}, a general-purpose instruction-following model selected to test whether strong language understanding capabilities transfer to medical dialogue tasks when combined with information-theoretic rewards, and DeepSeek-R1-Distill-Qwen-7B \cite{deepseekai2025deepseekr1incentivizingreasoningcapability}, a model optimized for reasoning tasks through reinforcement learning. We hypothesize that DeepSeek-R1-Distill-Qwen-7B's enhanced reasoning capabilities will better leverage our information gain rewards for strategic questioning, as evidenced by its larger relative improvements in our results (10.9\% on Avey, 12.9\% on MIMIC). We apply LoRA (rank=8) to both models for parameter-efficient fine-tuning.
We compare our IGFT-trained models against several baselines: the zero-shot performance of both base models without fine-tuning, GPT-4o-mini as a commercial baseline, and two medical domain-specific models, HuatuoGPT-o1-7B (trained on Chinese medical QA translated to English) \cite{wang2304huatuo} and UltraMedical-8B (pretrained on medical textbooks and literature) \cite{zhang2024ultramedical}.

We apply LoRA (rank=8) to reduce computational requirements while maintaining performance.

\subsection{Training Configuration}
We train models using a batch size of 2 per device with gradient accumulation steps of 4, resulting in an effective batch size of 8. The learning rate is set to 1e-4 with a constant warmup schedule. We use the AdamW optimizer with weight decay of 0.01. For sequence processing, we set a maximum length of 2048 tokens to accommodate multi-turn conversations. During GRPO training, we generate 2 responses per prompt for ranking-based optimization. Each training epoch consists of 10 steps, and we train for a total of 150 epochs to ensure convergence of the information gain rewards. 
We fine-tuned the models on a server with 1 NVIDIA A100 80GB GPU. The results are averaged over 3 runs with different random seeds to ensure reproducibility. Detailed parameter values and the ranges explored during development are provided in the Appendix.

\subsection{Evaluation Metrics}
We adopt the HPI evaluation framework from \cite{winston24medical}, which assesses the quality of AI-generated medical interviews by comparing atomic statements extracted from generated and ground truth HPIs.
Our evaluation process uses GPT-4o-mini to first extract atomic statements from both the generated and ground truth HPIs. GPT-4o-mini then determines semantic equivalence between these extracted statements, accounting for variations in phrasing while preserving clinical meaning. Based on this statement matching, we calculate precision and recall metrics as follows
\begin{align*}
\text{Recall} &= \frac{\text{matched ground truth statements}}{\text{total ground truth statements}}
\\
\text{Precision} &= \frac{\text{correct generated statements}}{\text{total generated statements}}   
\end{align*}

\begin{figure}[t] 
    \centering
    \includegraphics[scale=0.25]{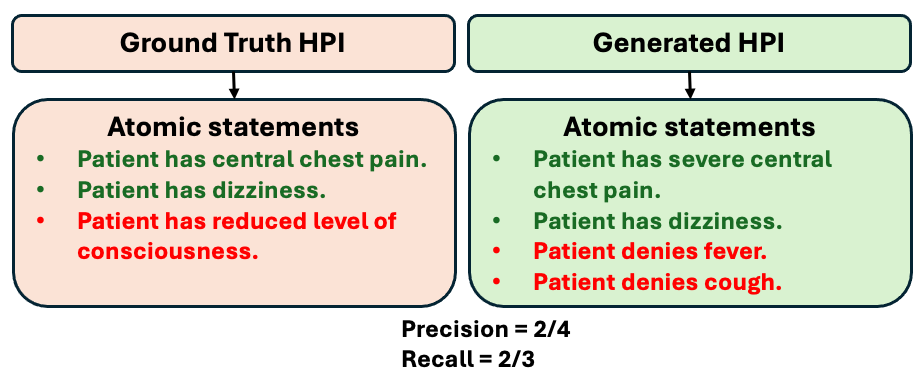}
        \caption{Example of HPI evaluation through atomic statement extraction and matching. Atomic statements are extracted from both ground truth and generated HPIs using GPT-4o-mini. Green indicates correctly captured information, while red indicates missed statements (left) or unnecessary information not present in the reference (right). In this example, the model correctly identified chest pain and dizziness but missed the reduced consciousness level and added denials of fever and cough, which were not part of the ground truth. This yields precision = 2/4 and recall = 2/3. }
    \label{fig:eval}
\end{figure}

Figure \ref{fig:eval} shows one example of evaluating HPI.

\section{Results}

\begin{table*}[htbp]
\centering
\begin{tabular}{|l|ccc|ccc|}
\toprule
&&\textbf{Avey}&&&\textbf{MIMIC}&\\
\textbf{Model} & 
\textbf{Precision} & 
\textbf{Recall} & 
\textbf{F1} & 
\textbf{Precision} & 
\textbf{Recall} & 
\textbf{F1} \\
\midrule
Llama-3.1-8B & 0.271$\pm$0.009 & 0.648$\pm$0.016 & 0.367$\pm$0.010 & 0.350$\pm$0.006 & 0.327$\pm$0.007 & 0.308$\pm$0.007 \\
Llama-3.1-8B (IGFT) & 0.286$\pm$0.014 & 0.671$\pm$0.014 & 0.384$\pm$0.013 & \textbf{0.378$\pm$0.023} & 0.340$\pm$0.019 & \textbf{0.336$\pm$0.018} \\
DeepSeek-R1-7B & 0.297$\pm$0.019 & 0.584$\pm$0.033 & 0.368$\pm$0.014 & 0.314$\pm$0.007 & 0.265$\pm$0.013 & 0.256$\pm$0.012 \\
DeepSeek-R1-7B (IGFT) & \textbf{0.329$\pm$0.003} & 0.611$\pm$0.017 & \textbf{0.408$\pm$0.005} & 0.348$\pm$0.011 & 0.280$\pm$0.015 & 0.289$\pm$0.018 \\
GPT-4o-mini & 0.268$\pm$0.009 & \textbf{0.687$\pm$0.036} & 0.372$\pm$0.003 & 0.325$\pm$0.008 & \textbf{0.344$\pm$0.017} & 0.304$\pm$0.001 \\
HuatuoGPT-o1-7B & 0.282$\pm$0.012 & 0.643$\pm$0.031 & 0.377$\pm$0.019 & 0.322$\pm$0.026 & 0.302$\pm$0.019 & 0.283$\pm$0.019 \\
UltraMedical-8B & 0.250$\pm$0.014 & 0.661$\pm$0.008 & 0.349$\pm$0.012 & 0.306$\pm$0.004 & 0.330$\pm$0.014 & 0.286$\pm$0.005 \\
\bottomrule
\end{tabular}
\caption{Model Performance on Avey and MIMIC Datasets. LLaMA-3.1-8B refers to Llama-3.1-8B-Instruct, and DeepSeek-R1-7B refers to DeepSeek-R1-Distill-Qwen-7B. Results show mean ± standard deviation across 3 runs.}
\label{tab:res}
\end{table*}

We evaluate IGFT approach on Llama-3.1-8B-Instruct and DeepSeek-R1-Distill-Qwen-7B. Models are trained exclusively on Avey data, which contains concise HPIs, yet we evaluate on both Avey and MIMIC datasets to assess generalization. The MIMIC dataset presents a more challenging test as it contains significantly longer and more elaborate HPIs. Table \ref{tab:res} presents precision, recall, and F1 scores the generated HPIs.
On the in-domain Avey test set, IGFT yields substantial improvements. DeepSeek-R1-Distill-Qwen-7B achieves an F1 score of 0.408±0.005, a 10.9\% relative improvement over its base model's 0.368±0.014, driven primarily by precision gains (0.329 vs 0.297). Llama-3.1-8B-Instruct improves to 0.384±0.013 from 0.367±0.010, with balanced improvements in both metrics. Both fine-tuned models surpass OpenAI's F1 score of 0.372±0.003, demonstrating that online learning with information gain rewards can match commercial systems without human conversation data.
The improvements transfer to the MIMIC dataset despite its different characteristics with longer, more detailed HPIs. DeepSeek-R1-Distill-Qwen-7B (IGFT) achieves 0.289±0.018 F1 versus the base model's 0.256±0.012, while Llama-8B (IGFT) reaches 0.336±0.018 compared to 0.308±0.007. The lower absolute scores on MIMIC reflect the increased complexity of generating comprehensive HPIs with many more clinical details. However, the consistent relative improvements suggest that information-theoretic rewards help models learn questioning strategies that scale to more complex cases. Notably, Llama-8B (IGFT) outperforms OpenAI's model on MIMIC (0.336 vs 0.304).
The reasoning-optimized DeepSeek-R1-Distill-Qwen-7B shows larger relative improvements than Llama-3.1-8B-Instruct on both datasets, suggesting that stronger reasoning capabilities help leverage information-theoretic rewards more effectively. The medical domain-specific models Huatuo-7B and UltraMedical-8B perform comparably to or below general-purpose base models despite their specialized training. This reflects a key distinction: these models were optimized for single-turn medical tasks, answering isolated questions or providing explanations, rather than multi-turn conversations requiring strategic information gathering. IGFT specifically addresses this gap by optimizing for sequential questioning strategies, demonstrating that effective medical history-taking requires different capabilities than traditional medical QA tasks.

Across all settings, IGFT consistently improves precision while maintaining or slightly improving recall. This precision-oriented behavior indicates that models learn to ask targeted questions for specific information rather than encouraging broad patient disclosure. This pattern holds even for the elaborate MIMIC HPIs, where precision improvements (e.g., 0.378 vs 0.350 for Llama-8B) suggest better focus on clinically relevant details despite the increased complexity. The small standard deviations demonstrate stable training despite the online RL framework.

\section{Discussion}

Our results validate the information-theoretic approach, with DeepSeek-R1-Distill-Qwen-7B achieving 10.9\% and 12.9\% F1 improvements on Avey and MIMIC respectively. Models trained on concise Avey vignettes generalized well to elaborate MIMIC HPIs, suggesting that information gain rewards help models learn fundamental principles of effective history-taking that transfer across documentation styles.

Several limitations warrant consideration. The computa-
tional cost of LLM-based reward computation remains high,
though reward model distillation could address this. Our
method currently requires pre-extracted entities from ground
truth HPIs; an end-to-end approach jointly learning entity
extraction and questioning would be more practical. While our approach handles multi-turn conversations, it optimizes each turn independently rather than planning across the entire dialogue, suggesting potential improvements through lookahead mechanisms. 

The framework extends beyond medical interviews to any
domain requiring strategic information gathering such as technical
support, legal interviews, educational assessment, or in-
vestigative journalism. This suggests objective-driven rein-
forcement learning may offer a broadly applicable paradigm
for training specialized AI systems.

Deploying AI in healthcare requires careful ethical con-
sideration. Patients must be informed when interacting with
AI systems, and AI-gathered histories should supplement,
not replace, physician judgment. Training data may reflect
healthcare disparities, necessitating evaluation across demo-
graphic groups to ensure equitable performance.

\section{Conclusion}

We presented Information Gain Fine-Tuning (IGFT), a novel approach for training medical conversational AI through online reinforcement learning with information-theoretic rewards. By formalizing medical history-taking as information acquisition, we eliminate dependence on scarce expert annotations while enabling models to discover effective questioning strategies through self-play.

IGFT-trained models consistently outperformed baselines and generalized from concise to elaborate HPIs, demonstrating that domain-specific objectives with quality constraints can guide models without extensive human supervision. While clinical validation remains necessary, this work provides a scalable path for developing AI tools to support more efficient patient care. Future work should address computational efficiency, end-to-end learning, and multi-turn strategic planning.

\bibliography{references}

\newpage


\section*{Supplementary material}

# Appendix: Aligning Medical Conversational AI through Online Reinforcement Learning with Information-Theoretic Rewards

## A  Detailed Hyperparameter Configuration

Both Llama-3.1-8B-Instruct and DeepSeek-R1-Distill-Qwen-7B models were fine-tuned using GRPO in an online reinforcement learning setup with identical hyperparameters except for reward scaling. The models used LoRA (rank 8, alpha 8) targeting attention projection layers [q_proj, k_proj, v_proj, o_proj] with no dropout. Training employed AdamW optimizer with fused operations (learning rate 1e-4, weight decay 0.01) and constant learning rate scheduling with 2 warmup steps that reset at the beginning of each online batch to ensure stable policy updates.

The effective batch size was 8 (per-device batch size 2 with gradient accumulation 4), appropriate for frequent policy updates in online RL. For GRPO configuration, both models generated 2 completions per prompt with maximum prompt and completion lengths of 2048 tokens each, using temperature $\tau$=1.0 for ranking. Generation parameters included temperature 0.7, top-p 0.9, top-k 50, and repetition penalty 1.2 to encourage diverse exploration during online learning.

Training followed a true online RL paradigm across 150 iterations (referred to as epochs). Each iteration used ground truth HPI from two patients to generate conversation episodes for training. The model was then fine-tuned on these episodes for 10 gradient steps, after which the updated model generated training data for the next iteration using the subsequent two patients' HPI. This online approach ensured the model continuously adapted its conversation generation strategy based on its evolving capabilities. The critical difference between models was in reward scaling: Llama-3.1-8B used no special scaling, while DeepSeek-R1-7B applied square root scaling (power 0.5) to IG scores and penalized low IG values (<0.01) by multiplying rewards by 0.1. We found that DeepSeek benefited from additional reward scaling due to its reasoning-optimized architecture producing different information gain distributions. Both models used $\lambda = 0.3$ for balancing information gain with quality assessment. The conversations during online training were generated for 10 turns ($N = 10$). For coverage probability estimation, we used $\alpha = 0.4, \beta = 0.3, \gamma = 0.3$, where these coefficients weight semantic similarity, relevance score, and keyword matching respectively.

We conducted systematic hyperparameter search across the following ranges: learning rate 5e-5, 1e-4, 5e-4, selecting 1e-4 for stable convergence. We selected LoRA rank 8 as higher ranks resulted in GPU out-of-memory issues.

## B  System Prompts

### Patient Simulation Prompt

The patient simulation prompt is designed to create realistic, gradual information revelation:

Listing 1: Complete patient simulation instructions

```
You are acting as a patient with the following medical characteristics:

Age: {age}
Sex: {sex}

History of Present Illness:
{hpi_first_person}

You DON'T have any of these symptoms:
{absent_findings_list}

Past Medical & Surgical History:
{past_medical_history}
```

```
Family History:
{family_history}

When answering the doctor's questions:
1. DO NOT list all your symptoms at once - reveal them gradually
   as the doctor asks about them
2. Answer ONLY the specific questions asked by the doctor
3. If asked about a symptom that isn't mentioned in your
   information, say you don't have it
4. Don't invent new symptoms or medical history beyond what's listed
5. Speak naturally as a patient would, in first person, without
   medical jargon
6. Show appropriate concern about your symptoms, but don't
   diagnose yourself
7. If asked about the duration of symptoms, use the timeframe
   mentioned in your history
8. Keep your answers brief and to the point - 1-3 sentences maximum
9. Only mention related symptoms when they're directly relevant
   to what the doctor just asked

Your goal is to realistically portray a patient with these
specific medical characteristics, revealing information gradually
as a real patient would.
```

## Doctor Interview Prompt

The doctor prompt enforces JSON output format and telehealth constraints:

Listing 2: Doctor interview instructions with JSON enforcement

```
You are a skilled physician conducting a TELEHEALTH patient interview.
Your goal is to determine the diagnosis through careful questioning
and diagnostic reasoning.

## Output Format Requirements
You MUST respond with a valid JSON object and nothing else.
Follow these rules exactly:

1. Your response must start with an opening curly brace '{'
   and end with a closing curly brace '}'
2. Use the following JSON structure:
{
    "doctor": "Your question to the patient",
    "doctor_thought": "Your medical reasoning"
}

3. Do NOT include any text outside the JSON object
4. Do NOT use triple backticks or markdown code blocks
5. Ensure all property names have double quotes
6. Ensure all string values have double quotes
7. Make sure all objects have proper comma separation between properties
8. Do not include trailing commas after the last property
9. Escape any quotes inside strings with a backslash

This format is mandatory for our medical system to process
your responses correctly.

You know only the following about this patient initially:
- Age: {age}
- Sex: {sex}

**IMPORTANT INSTRUCTIONS:**
1. This is a TELEHEALTH encounter - you cannot directly
   examine the patient.
2. You CAN ask the patient about self-reported vital signs
   (temperature, blood pressure, heart rate, oxygen levels)
```

```
      if they have home monitoring devices.
   3. The patient can only report symptoms they can observe
      or feel themselves.
   4. Do not include any hands-on physical examination
      (like auscultation, palpation, etc.).
   5. Start with open-ended questions, then move to more
      specific ones as needed.
   6. Ask questions one at a time - patients will only provide
      information about what you specifically ask.
   7. Ensure the diagnostic hypotheses evolve naturally as new
      information is revealed.

Always ask only ONE question at a time - this is a realistic
medical interview.

Begin the interview by asking about the patient's chief complaint.
```

## C  Example Outputs and Qualitative Analysis

Table C provides an example of a generated conversation by Llama-3.1-8B (IGFT), while Table 1 presents the generated HPI alongside the corresponding ground truth HPI from the Avey dataset. Table 2 provides a detailed analysis of the extracted statements. The generated HPI demonstrates perfect recall (100%), indicating that all ground truth statements were successfully captured. However, the precision is relatively low (23.1%) due to the inclusion of extensive additional information not present in the ground truth. Much of this additional content consists of detailed clinical descriptions and inferences that, while not explicitly stated in the reference HPI, are valid clinical observations derived from the conversation. This suggests that the model is capable of extracting relevant information but may also overgenerate by incorporating clinically plausible, yet unsupported, details.

| Turn | Speaker | Conversation |
|---|---|---|
| 1 | **Doctor:** | Hello, I'm Dr. Buddy. What brings you in today? |
|  | **Patient:** | I'm here because of fatigue and paresthesia. |
| 2 | **Doctor:** | Can you tell me more about when these symptoms started? Was there anything that triggered them? |
|  | **Patient:** | The symptoms started about eight months ago, and I noticed they began happening after I started a strict vegan diet about a year ago. I'm not sure if that's related, but it seems like it might be. |
| 3 | **Doctor:** | Can you describe the nature of the paresthesia you're experiencing? |
|  | **Patient:** | The paresthesia feels like tingling and numbness, mostly in my hands and feet. It's been pretty bothersome, especially when I'm trying to do tasks. |
| 4 | **Doctor:** | Have you experienced any changes in your bowel movements since starting the vegan diet? |
|  | **Patient:** | I haven't really noticed any significant changes in my bowel movements since starting the vegan diet. Everything seems normal in that regard. |
| 5 | **Doctor:** | Can you estimate how long these episodes of tingling and numbness typically last? |
|  | **Patient:** | The tingling and numbness episodes usually last for a few minutes, but they can sometimes linger for longer, especially if I'm doing something that involves my hands or feet. |
| 6 | **Doctor:** | How would you describe the intensity of these episodes of tingling and numbness on a scale from 0 to 10? |
|  | **Patient:** | I would say the intensity of the tingling and numbness is around a 6 out of 10. It can be quite uncomfortable, especially when it happens frequently. |
| 7 | **Doctor:** | Have you had any recent weight loss or unintentional weight gain? |
|  | **Patient:** | I haven't noticed any recent weight loss or gain. My weight has been pretty stable. |
| 8 | **Doctor:** | Have you noticed any other neurological symptoms such as weakness, difficulty walking, blurred vision, or seizures? |
|  | **Patient:** | Yes, I've been experiencing weakness and an unsteady gait as well. It's been concerning, especially since I've had these symptoms for the same duration as the others, around eight months. |
| 9 | **Doctor:** | Can you also let me know whether these neurological symptoms worsen at night while resting, or if they tend to improve throughout the day? |
|  | **Patient:** | I've noticed that the weakness and unsteady gait tend to worsen at night while I'm resting. It's frustrating because I feel like I can't relax without feeling off balance. |
| 10 | **Doctor:** | Can you think of any potential triggers or factors that may have contributed to these symptoms worsening at night? |
|  | **Patient:** | I haven't really identified any specific triggers for the symptoms worsening at night. It just seems to happen, and I can't pinpoint what might be causing it. |

| Generated HPI | Ground Truth HPI |
|---|---|
| The patient is a 55-year-old female presenting with an eight-month history of fatigue and neurological symptoms, including paresthesia, weakness, and an unsteady gait. She reports that her symptoms began approximately eight months ago, a few months after transitioning to a strict vegan diet, raising concern for possible nutritional deficiencies. The paresthesias are described as episodes of tingling and numbness that primarily affect her hands and feet. These episodes typically last a few minutes but can persist for longer durations, especially with activities that require the use of her hands and feet, and are rated about 6 out of 10 in intensity. She denies any significant changes in bowel habits, noting that her bowel movements have remained normal since the diet change, and she has not experienced notable weight loss or gain. The patient also describes a worsening of the weakness and unsteady gait at night while resting, which disrupts her ability to relax and sleep. There are no specific triggers she can identify for the exacerbation of her nighttime symptoms. Given the temporal association with the dietary change and the pattern of neurological deficits, there is a concern for a possible nutritional deficiency, such as vitamin B12 deficiency, which is known to cause similar symptoms. Overall, her clinical picture is marked by progressive neurological deficits that are affecting her daily functional abilities, warranting further laboratory investigations to assess her nutritional status and elucidate the underlying etiology of her symptomatology. | Age: 55 years. Sex: Female, Chief complaints: Fatigue and Paresthesia History of presenting illness: Comes to your office with an 8-month history of increasing fatigue, paresthesias, weakness, and an unsteady gait. These are new symptoms that started to appear after she began her strict vegan diet a year ago. Absent findings: No fever or chills, No vaginal bleeding, No melena or hematochezia, No urinary symptoms, No dysphagia, No anhedonia, No mood swings, No oral ulcers, No motor weakness, No headache, No ear pain, No hearing loss, No bone pain, No cold or heat intolerance, No change in bowel habits. Physical examination: Normal vitals. Her skin is pale, as are her lower conjunctival lids. Neurological examination: Patchy impairment of the sensations of touch and temperature, loss of vibration and position sense. Past medical and surgical history: Diabetes Mellitus 2 years ago, controlled on medications. |

Table 1: Example of generated HPI and corresponding ground-truth HPI.

| Statement | Type | Supported |
|---|---|---|
| **Ground Truth Statements (Reference)** | | |
| Patient has an 8-month history of increasing fatigue | GT | ✓ |
| Patient has paresthesias | GT | ✓ |
| Patient has weakness | GT | ✓ |
| Patient has an unsteady gait | GT | ✓ |
| Symptoms are new | GT | ✓ |
| Symptoms started to appear after she began her strict vegan diet a year ago | GT | ✓ |
| **Generated Statements (Correctly Captured)** | | |
| Patient has an eight-month history of fatigue | Gen | ✓ |
| Patient experiences paresthesia | Gen | ✓ |
| Patient experiences weakness | Gen | ✓ |
| Patient has an unsteady gait | Gen | ✓ |
| Symptoms began approximately eight months ago | Gen | ✓ |
| Symptoms began a few months after transitioning to a strict vegan diet | Gen | ✓ |
| **Generated Statements (Additional/Unsupported)** | | |
| Patient has neurological symptoms | Gen | ✗ |
| There is concern for possible nutritional deficiencies | Gen | ✗ |
| Paresthesias are described as episodes of tingling and numbness | Gen | ✗ |
| Paresthesias primarily affect her hands and feet | Gen | ✗ |
| Paresthesia episodes typically last a few minutes | Gen | ✗ |
| Paresthesia episodes can persist for longer durations | Gen | ✗ |
| Paresthesia episodes are exacerbated by activities... | Gen | ✗ |
| Paresthesia intensity is rated about 6 out of 10 | Gen | ✗ |
| Patient denies any significant changes in bowel habits | Gen | ✗ |
| Patient's bowel movements have remained normal since the diet change | Gen | ✗ |
| Patient has not experienced notable weight loss | Gen | ✗ |
| Patient has not experienced notable weight gain | Gen | ✗ |
| Patient describes worsening weakness at night while resting | Gen | ✗ |
| Patient describes worsening unsteady gait at night while resting | Gen | ✗ |
| Worsening symptoms disrupt her ability to relax and sleep | Gen | ✗ |
| There are no specific triggers identified for nighttime symptoms | Gen | ✗ |
| There is concern for a possible nutritional deficiency (B12) | Gen | ✗ |
| Patient's clinical picture is marked by progressive neurological deficits | Gen | ✗ |
| Progressive neurological deficits are affecting daily functional abilities | Gen | ✗ |
| Further laboratory investigations are warranted. | Gen | ✗ |

Table 2: Statement level analysis.

| Metric | Value |
| --- | --- |
| Precision | 0.231 (23.1%) |
| Recall | 1.000 (100.0%) |
| F1 Score | 0.375 (37.5%) |
| Supported Ground Truth Statements | 6/6 |
| Supported Generated Statements | 6/26 |

Table 3: Evaluation Metric


\end{document}